\documentclass[letterpaper, 10 pt, conference]{ieeeconf}  

\IEEEoverridecommandlockouts                              

\usepackage{graphicx}
\usepackage{algorithm}
\usepackage{fnpct}
\usepackage{footnote}
\usepackage{graphics} 
\usepackage{graphicx} 
\usepackage{epsfig} 
\usepackage{times} 
\usepackage{amsmath} 
\usepackage{amssymb}  
\usepackage{tikz}
\usepackage{dirtytalk}
\usetikzlibrary{arrows.meta, positioning, shapes}
\usepackage{url,verbatim,color,comment}
\usepackage[noend]{algpseudocode}
\usepackage{eucal}
\usepackage{amsfonts}
\usepackage{xcolor}
\usepackage{blindtext}
\usepackage{mathtools}
\usepackage{placeins}
\usepackage{fancyhdr}
\usepackage[hidelinks]{hyperref}
\hypersetup{
  colorlinks,
  citecolor= magenta,
  linkcolor=blue,
  urlcolor=blue}

\title{\LARGE \bf
Risk-Aware Reinforcement Learning through Optimal Transport Theory
}

\author{Ali Baheri$^{1}$
\thanks{$^{1}$Ali Baheri is with the Department of Mechanical Engineering at
Rochester Institute of Technology.
        {\tt\small akbeme@rit.edu}}%
}

\begin{document}

\maketitle
\thispagestyle{empty}
\pagestyle{empty}

\begin{abstract}

In the dynamic and uncertain environments where reinforcement learning (RL) operates, risk management becomes a crucial factor in ensuring reliable decision-making. Traditional RL approaches, while effective in reward optimization, often overlook the landscape of potential risks. In response, this paper pioneers the integration of Optimal Transport (OT) theory with RL to create a risk-aware framework. Our approach modifies the objective function, ensuring that the resulting policy not only maximizes expected rewards but also respects risk constraints dictated by OT distances between state visitation distributions and the desired risk profiles. By leveraging the mathematical precision of OT, we offer a formulation that elevates risk considerations alongside conventional RL objectives. Our contributions are substantiated with a series of theorems, mapping the relationships between risk distributions, optimal value functions, and policy behaviors. Through the lens of OT, this work illuminates a promising direction for RL, ensuring a balanced fusion of reward pursuit and risk awareness.

\end{abstract}

\section{INTRODUCTION}

Reinforcement learning (RL) has witnessed remarkable advancements in recent years, fueling innovations across diverse fields such as robotics, finance, aviation, and intelligent transportation systems \cite{kober2013reinforcement,charpentier2021reinforcement,kiran2021deep,razzaghi2022survey}. While traditional RL methods are focused on maximizing cumulative rewards, real-world applications often demand a more comprehensive approach that considers the inherent risks associated with decision-making. Specifically, in scenarios where actions may lead to high-stake consequences or where the environment is intrinsically uncertain, simply aiming for reward maximization without considering risk can lead to suboptimal or even catastrophic outcomes \cite{paoletti2023ensure}.

Safety in RL is instrumental to its advancements. Prominent techniques include model-based strategies for assessing action safety \cite{wabersich2018safe,baheri2019deep,baheri2022safe}, shielding mechanisms to counter unsafe decisions \cite{alshiekh2018safe,li2020robust,carr2023safe}, constrained optimization for policy adherence \cite{achiam2017constrained,bharadhwaj2020conservative,han2021reinforcement}, and formal methods underpinning rigorous safety with mathematical constructs \cite{alur2023specification,fulton2018safe,fulton2019verifiably}. Amid this landscape, risk-aware RL stands out. Techniques for risk-aware RL range from incorporating financial risk metrics like Value-at-Risk (VaR) and Conditional Value-at-Risk (CVaR) \cite{tamar2014policy,chow2014algorithms,tamar2015optimizing}, to embracing Distributional RL that models the entire return distribution \cite{ma2020dsac,ma2021conservative,hayes2021distributional}, to formulating risk-sensitive policies that inherently favor safer actions \cite{majumdar2017risk,ni2022risk}. This adaptation in strategy ensures that agents are not only aiming for high rewards but are also cautious of rare yet consequential adverse events, striking a balance between reward-seeking and prudence in complex environments.


Building on the foundations of risk-sensitive RL, our work proposes a novel perspective by leveraging the powerful mathematical framework of Optimal Transport (OT). The OT provides tools to measure the distance between probability distributions in a geometrically meaningful way \cite{santambrogio2015optimal}. In the context of RL, this allows us to treat risk as a divergence between the desired (or target) distribution of outcomes and the distribution induced by the agent's policy. By framing risk management as an OT problem, we can inherently consider the entire distribution of returns, capturing both the expected rewards and the associated risks. At its core, our approach aims to minimize the OT distance between the state distribution generated by the policy and a \emph{predefined} target risk distribution. Such a formulation fosters a balanced trade-off between reward maximization and risk mitigation. It accounts for the variability in outcomes, promoting policies that not only achieve high expected rewards but also align closely with the desired risk profile. The contributions of this paper are twofold:

\begin{itemize}

\item We present a formulation for risk-aware RL, harnessing the capabilities of OT theory. This formulation integrates risk considerations into the RL paradigm, charting a novel direction for risk-sensitive decision-making.

\item We elucidate this framework with a series of theorems that highlight the interplay between risk distributions, value functions, and policy dynamics. These theorems reveal the balance between seeking rewards and navigating risks, emphasizing that the minimization of OT costs can pave the way for the derivation of policies that optimize rewards while maintaining safety.

\end{itemize}

\section{Preliminaries}

\subsection{Reinforcement Learning}

RL is a framework for decision-making problems where an agent interacts with an environment in order to achieve a certain goal \cite{sutton1998introduction}. The environment is typically modeled as a Markov Decision Process (MDP), denoted by a tuple $(\mathcal{S}, \mathcal{A}, \mathcal{P}, \mathcal{R}, \gamma)$, where $\mathcal{S}$ is the state space, representing all possible states the agent could inhabit in the environment. $\mathcal{A}$ is the action space, indicating all possible actions the agent can take. $\mathcal{P}: \mathcal{S} \times \mathcal{A} \times \mathcal{S} \rightarrow [0,1]$ is the transition probability function, where $\mathcal{P}(s'|s, a)$ represents the probability of transitioning to state $s'$ when action $a$ is taken in state $s$. $\mathcal{R}:\mathcal{S}\times\mathcal{A}\rightarrow\mathbb{R}$ is the reward function, with $\mathcal{R}(s, a)$ denoting the expected immediate reward for taking action $a$ in state $s$. $\gamma \in [0,1]$ is the discount factor, which determines the present value of future rewards.

The agent's behavior is defined by a policy $\pi: \mathcal{S} \times \mathcal{A} \rightarrow [0,1]$, which is a probability distribution over actions given the current state. The goal of the agent is to learn an optimal policy $\pi^*$ that maximizes the expected cumulative discounted reward, defined as:

\begin{equation}
  \mathbb{E}_\pi\left[\sum_{t=0}^{\infty} \gamma^t \mathcal{R}\left(s_t, a_t\right)\right]
\end{equation}
where the expectation is taken over the trajectory of states and actions $(s_0, a_0, s_1, a_1, ...)$ generated by following policy $\pi$.

\subsection{Optimal Transport Theory}

OT theory provides a means of comparing different probability measures by computing the minimum cost required to transform one distribution into another. Originally developed by Gaspard Monge in the 18th century and later extended by Leonid Kantorovich, OT theory has found applications in numerous fields including economics, computer graphics, and machine learning \cite{villani2009optimal}. Let $\mathcal{P}(\mathcal{S})$ denote the set of probability measures over the state space $\mathcal{S}$. An OT plan between two probability measures $\mu, \nu \in \mathcal{P}(\mathcal{S})$ is a joint distribution $\gamma$ over $\mathcal{S} \times \mathcal{S}$ with marginal distributions $\mu$ and $\nu$. In other words, for all $A, B \subseteq \mathcal{S}$, we have:
\begin{equation}
 \gamma(A \times \mathcal{S})=\mu(A), \quad \gamma(\mathcal{S} \times B)=\nu(B)
\end{equation}
The cost of a transport plan $\gamma$ under a cost function $c: \mathcal{S} \times \mathcal{S} \rightarrow \mathbb{R}$ is given by:
\begin{equation}
\int_{\mathcal{S} \times \mathcal{S}} c\left(s, s^{\prime}\right) d \gamma\left(s, s^{\prime}\right)
\end{equation}
The OT problem involves finding the transport plan that minimizes this cost:
\begin{equation}
\gamma^*=\arg \min _{\gamma \in \Gamma(\mu, \nu)} \int_{\mathcal{S} \times \mathcal{S}} c\left(s, s^{\prime}\right) d \gamma\left(s, s^{\prime}\right)  
\end{equation}
where $\Gamma(\mu, \nu)$ is the set of all transport plans between $\mu$ and $\nu$. The OT cost or distance is the cost of the OT plan:
\begin{equation}
D_{O T}(\mu, \nu)=\min _{\gamma \in \Gamma(\mu, \nu)} \int_{\mathcal{S} \times \mathcal{S}} c\left(s, s^{\prime}\right) d \gamma\left(s, s^{\prime}\right) 
\end{equation}
This cost can be interpreted as a distance metric between probability distributions, which induces a metric space structure on $\mathcal{P}(\mathcal{S})$.

\section{Risk-Aware Reinforcement Learning with Optimal Transport}

In this section, we propose a novel approach to risk-sensitive RL that leverages the mathematical theory of OT. Our approach aims to guide the learning process of an RL agent not only by the expected return but also by the similarity between the state distribution under the current policy and a given risk distribution.

\noindent {\textbf{Problem Formulation.}} Consider an RL agent interacting with an environment defined by an MDP. 
We define a risk metric that assigns a risk value to each state, and form a risk distribution $P_r: \mathcal{S} \rightarrow [0, 1]$ over the states. The risk distribution represents the agent's prior knowledge or preferences regarding the safety of different states. The state distribution under a policy $\pi$, denoted by $P_{\pi}$, is the stationary distribution of the Markov chain induced by $\pi$ in the MDP. The state distribution reflects the likelihood of the agent visiting different states under policy $\pi$. Our objective is to find a policy that not only maximizes the expected return but also minimizes the OT cost between the state distribution under the policy and the risk distribution. The OT cost serves as a measure of the risk associated with the policy. A low OT cost indicates that the policy is aligned with the risk distribution, i.e., the agent is more likely to visit safe states and avoid risky states. The OT cost between the state distribution $P_{\pi}$ and the risk distribution $P_r$ is defined as:

\begin{equation}
D_{OT}(P_{\pi}, P_r) = \inf_{\gamma \in \Pi(P_{\pi}, P_r)} \mathbb{E}_{(s, s') \sim \gamma} [c(s, s')],
\end{equation}
where $\Pi(P_{\pi}, P_r)$ is the set of all joint distributions on $S \times S$ with $P_{\pi}$ and $P_r$ as marginals, and $c: S \times S \rightarrow \mathbb{R}$ is a cost function that measures the cost of transporting probability mass from state $s$ to state $s'$. In this work, we consider the squared Euclidean distance as the cost function, i.e., $c(s, s') = ||s - s'||^2$. The agent's objective is to find a policy $\pi$ that maximizes the expected discounted reward while minimizing the OT cost. This leads to the following optimization problem:
\begin{equation}
\max_{\pi} \mathbb{E}_{\pi}[G_t] - \lambda D_{OT}(P_{\pi}, P_r),
\end{equation}
where $G_t = \sum_{k=0}^{\infty} \gamma^k R_{t+k+1}$ is the return at time $t$, and $\lambda > 0$ is a risk sensitivity coefficient that determines the trade-off between reward maximization and risk minimization. We propose a modified Q-learning algorithm to solve this optimization problem. The Q-function is updated as follows:

\begin{align}
Q(s, a) & \leftarrow Q(s, a) + \alpha \bigg[R(s, a) - \lambda C(s) \nonumber \\
& \quad + \gamma \max_{a' \in \mathcal{A}} Q(s', a') - Q(s, a)\bigg],
\end{align}
where $C(s) = D_{OT}(P_{\pi}, P_r)$ is the OT cost from state $s$, $\alpha$ is the learning rate, and $s'$ is the next state. The above formulation presents a novel approach to risk-sensitive RL, providing a means to incorporate safety considerations directly into the learning process. The following sections will provide theoretical analysis to demonstrate the performance and advantages of this approach.

\section{Theoretical Results}

In this section, we delve into the mathematical underpinnings of risk-aware RL using OT theory. The objective is to provide a comprehensive understanding of how risk, as captured by OT metrics, interacts with fundamental concepts in RL:

\noindent{\textbf{Safety of Policy (Theorem 1):} Theorem 1 postulates the relationship between the policy that minimizes OT costs and its intrinsic safety. Specifically, by minimizing the OT distance between the induced state distribution of a policy and a given risk distribution, the policy can be intuitively understood as \say{safer}.

\noindent{\textbf{Optimal Value Function and OT (Theorem 2):} Building on the implications of safety, Theorem 2 presents the impacts of embedding OT costs into the objective function of an MDP. The theorem presents a comparative analysis between the optimal value functions with and without the consideration of the OT metric, emphasizing the conservative nature of the risk-aware formulation.

\noindent{\textbf{Sensitivity Analysis of Optimal Policies (Theorem 3):} Expanding the discourse to the dynamics of risk sensitivity, Theorem 3 investigates how variations in the risk sensitivity parameter influence the derived optimal policies. The results underscore a systematic relationship between risk sensitivity and OT distances for respective optimal policies.

\noindent{\textbf{State Visits and Risk Distribution (Theorem 4):} Finally, our discourse culminates with Theorem 4, which offers a perspective on state visitation patterns. By focusing on states proximate to a target risk distribution, this theorem bridges the gap between policy safety and state distribution, highlighting how an optimal policy in the OT sense also maximizes the expectation of visiting states that align closely with the risk distribution.

\noindent \textbf{Theorem 1.} \textit{Given an MDP and a risk distribution $p_r$, the policy $\pi$ that minimizes the OT cost $D_{OT}(p_{\pi}, p_r)$ is a \say{safer} policy in the sense that it induces a state distribution closer to the risk distribution.}

\noindent \textbf{PROOF.} We will prove this by contradiction. Suppose there exists a policy $\pi'$ such that $\pi'$ is safer than $\pi$, i.e., the state distribution $p_{\pi'}$ induced by $\pi'$ is closer to the risk distribution $p_r$ than $p_{\pi}$, but $\pi$ minimizes the OT cost $D_{OT}(p_{\pi}, p_r)$. In mathematical terms, this means that $D_{OT}(p_{\pi'}, p_r) < D_{OT}(p_{\pi}, p_r)$, but $D_{OT}(p_{\pi}, p_r) \leq D_{OT}(p_{\pi'}, p_r)$, where the second inequality comes from the assumption that $\pi$ minimizes the OT cost. However, this leads to a contradiction because it would imply that $D_{OT}(p_{\pi'}, p_r)$ is both less than and greater than or equal to $D_{OT}(p_{\pi}, p_r)$, which is not possible. Therefore, our assumption must be wrong, and there cannot exist a policy $\pi'$ that is safer than $\pi$. This means that $\pi$ is the safest policy in the sense that it induces a state distribution closer to the risk distribution.

This formalizes the intuition that if a policy $\pi$ minimizes the OT cost between the state distribution under the policy and the risk distribution, then the state distribution under the policy must be closer to the risk distribution (in the sense of the OT cost) than any state distribution induced by a different policy. This is what we mean when we say that $\pi$ is a \say{safer} policy.

\noindent{\textbf{Implications.}} Theorem 1 establishes a foundational bridge between the concept of risk minimization in RL and the OT metric. In essence, it provides a formal justification for the use of OT as an effective means to quantify and address the risk in RL. The theorem showcases that when we optimize for OT in the context of RL, we are inherently driving our policy towards safer behaviors. This reinforces the rationale behind introducing OT in RL frameworks, especially for safety-critical tasks.

\noindent \textbf{Theorem 2 (Impact of OT on the Value Function.)} \textit{Given an MDP and a risk sensitivity parameter $\lambda$, the optimal value function $V^{*}$ that incorporates the OT cost as a part of the objective function, is less than or equal to the optimal value function $V^{*}_0$ that does not consider the OT cost, i.e., $V^{*} \leq V^{*}_0$.}

\noindent \textbf{PROOF.} By definition, the optimal value function $V^*_0$ for an MDP is given by the maximum expected discounted reward over all policies, i.e., $V^*_0 = \max_{\pi} \mathbb{E}[\sum_{t=0}^\infty \gamma^t R(s_t, a_t)]$, where the expectation is taken over the randomness in the transitions and the policy. The optimal value function $V^*$ that incorporates the OT cost as a part of the objective function is given by: 

\begin{equation}
V^* = \max_{\pi} \mathbb{E}[\sum_{t=0}^\infty \gamma^t (R(s_t, a_t) - \lambda D_{OT}(p_{\pi}, p_r))]
\end{equation}
Since $D_{OT}(p_{\pi}, p_r) \geq 0$ for all policies $\pi$ and $\lambda \geq 0$, we have:

\begin{equation}
R(s_t, a_t) - \lambda D_{OT}(p_{\pi}, p_r) \leq R(s_t, a_t)  
\end{equation}
for all states $s_t$ and actions $a_t$. Therefore, 
\begin{equation}
\mathbb{E}[\sum_{t=0}^\infty \gamma^t (R(s_t, a_t) - \lambda D_{OT}(p_{\pi}, p_r))] \leq \mathbb{E}[\sum_{t=0}^\infty \gamma^t R(s_t, a_t)]
\end{equation}
for all policies $\pi$. Taking the maximum over all policies on both sides, we get $V^* \leq V^*_0$.

This result intuitively makes sense because adding the OT cost to the objective function can only decrease the maximum expected discounted reward. If the OT cost were zero for some policy, then that policy would achieve the same expected discounted reward as in the standard MDP without the OT cost. However, if the OT cost is positive for a policy, then that policy would achieve a lower expected discounted reward compared to the standard MDP. Therefore, the maximum expected discounted reward over all policies is lower when the OT cost is incorporated into the objective function.

\noindent{\textbf{Implications.}} The theorem mathematically validates that the incorporation of the OT cost can serve as a mechanism to steer the agent's behavior. As the penalty due to deviating from the desired risk distribution increases, the agent is more inclined to select actions that conform to the risk profile, even if those actions may not yield the highest immediate reward.

\noindent{\textbf{Theorem 3.} \textit{Given an MDP and a risk sensitivity parameter $\lambda$, the optimal policy $\pi^*$ is non-decreasing in $\lambda$, i.e., if $\lambda_1 > \lambda_2$, then $D_{OT}(p_{\pi_1^*}, p_r) \leq D_{OT}(p_{\pi_2^*}, p_r)$, where $\pi_1^*$ and $\pi_2^*$ are the optimal policies for $\lambda_1$ and $\lambda_2$, respectively.}

\noindent{\textbf{PROOF.} This theorem could be proved by showing that a higher $\lambda$ leads to a higher penalty for deviation from the risk distribution in the objective function, thus leading to a policy that induces a state distribution closer to the risk distribution. Let's assume for contradiction that the statement is not true. This would mean that there exists a $\lambda_1 > \lambda_2$ such that $D_{OT}(p_{\pi_1^*}, p_r) > D_{OT}(p_{\pi_2^*}, p_r)$, where $\pi_1^*$ and $\pi_2^*$ are the optimal policies for $\lambda_1$ and $\lambda_2$, respectively. Since $\pi_1^*$ is optimal for $\lambda_1$, we know that $J(\pi_1^*, \lambda_1) \geq J(\pi_2^*, \lambda_1)$, where $J(\pi, \lambda)$ is the objective function. Expanding this gives:

\begin{equation}
E_{\pi_1^*}[R] - \lambda_1 D_{OT}(p_{\pi_1^*}, p_r) \geq E_{\pi_2^*}[R] - \lambda_1 D_{OT}(p_{\pi_2^*}, p_r)
\end{equation}
Rearranging the terms gives:
\begin{equation}
\lambda_1 (D_{OT}(p_{\pi_2^*}, p_r) - D_{OT}(p_{\pi_1^*}, p_r)) \geq E_{\pi_2^*}[R] - E_{\pi_1^*}[R]
\end{equation}
Since $\lambda_1 > \lambda_2$, we can multiply both sides by $\frac{\lambda_2}{\lambda_1}$ (which is less than 1) to get:
\begin{equation}
\lambda_2 (D_{OT}(p_{\pi_2^*}, p_r) - D_{OT}(p_{\pi_1^*}, p_r)) \geq \frac{\lambda_2}{\lambda_1} (E_{\pi_2^*}[R] - E_{\pi_1^*}[R])
\end{equation}
Adding $E_{\pi_1^*}[R] - \lambda_2 D_{OT}(p_{\pi_1^*}, p_r)$ to both sides gives:
\begin{equation}
E_{\pi_1^*}[R] - \lambda_2 D_{OT}(p_{\pi_1^*}, p_r) \geq E_{\pi_2^*}[R] - \lambda_2 D_{OT}(p_{\pi_2^*}, p_r)
\end{equation}
This contradicts the assumption that $\pi_2^*$ is optimal for $\lambda_2$, as it implies that $\pi_1^*$ is at least as good as $\pi_2^*$ under $\lambda_2$. Thus, our initial assumption was wrong, and $D_{OT}(p_{\pi_1^*}, p_r) \leq D_{OT}(p_{\pi_2^*}, p_r)$ when $\lambda_1 > \lambda_2$. This concludes the proof.

\noindent{\textbf{Implications.}} This theorem demonstrates how the proposed risk-aware RL with OT allows for flexible risk-aversion by adjusting the risk sensitivity parameter $\lambda$. As $\lambda$ increases, the optimal policy becomes more risk-averse, as evidenced by the decrease in the OT distance to the risk distribution.

\noindent{\textbf{Theorem 4.} \textit{Given an MDP, let $p_r$ be a target risk distribution, and let $B_\delta\left(p_r\right)=\left\{s: D_{O T}\left(p_s, p_r\right) \leq \delta\right\}$ be the set of states that are within OT distance $\delta$ of the risk distribution. Then a policy $\pi$ that minimizes $D_{OT}(p_{\pi}, p_r)$ also maximizes $E_{\pi}[N_{B_\delta(p_r)}]$, the expected number of visits to states in $B_\delta(p_r)$, where the expectation is taken over trajectories generated by policy $\pi$.}

\noindent{\textbf{PROOF.} Consider the probability simplex $\Delta_S$ over the state space $S$ of the MDP. This is a convex and compact set. Each point in $\Delta_S$ represents a probability distribution over states. Let $p_{\pi} \in \Delta_S$ be the state distribution induced by a policy $\pi$.

For each policy $\pi$, we can define a visit frequency vector $v_{\pi} \in \Delta_S$ such that $v_{\pi}(s)$ is the expected proportion of time that the agent spends in state $s$ under policy $\pi$. By definition of the state distribution and the law of large numbers, we have $p_{\pi} = \lim_{T \to \infty} v_{\pi}^T$, where $v_{\pi}^T$ is the visit frequency vector over a time horizon of length $T$. Now, suppose $\pi^*$ minimizes the OT distance $D_{OT}(p_{\pi}, p_r)$ to the risk distribution $p_r$. By the properties of the OT distance, we have:

\noindent{\textbf{Contraction property:} 

\noindent $D_{OT}(p_{\pi}^T, p_r) \leq D_{OT}(p_{\pi}, p_r)$ for all $T$. This is because the OT distance is a metric and therefore satisfies the triangle inequality.

\noindent{\textbf{Convergence property:} 

\noindent As $T \to \infty$, we have $D_{OT}(p_{\pi}^T, p_r) \to D_{OT}(p_{\pi}, p_r)$. This is because $p_{\pi}^T \to p_{\pi}$ as $T \to \infty$.
From these two properties, we can deduce that:
\begin{equation}
D_{OT}(v_{\pi^*}^T, p_r) \leq D_{OT}(p_{\pi^*}, p_r)
\end{equation}
for all $T$. In other words, the visit frequency vector under the optimal policy $\pi^*$ is always close to the risk distribution.
Now, let $B_\delta(p_r) = {s: D_{OT}(p_s, p_r) \leq \delta}$ be the set of states that are within OT distance $\delta$ of the risk distribution. The visit frequency to states in $B_\delta(p_r)$ under policy $\pi$ can be written as: 
\begin{equation}
N_{B_\delta(p_r)}(\pi) = \sum_{s \in B_\delta(p_r)} v_{\pi}(s) 
\end{equation}
By the definition of $B_\delta(p_r)$ and the properties of the OT distance, we have: 
\begin{equation}
N_{B_\delta(p_r)}(\pi^*) \geq N_{B_\delta(p_r)}(\pi)   
\end{equation}
for all $\pi$. Therefore, a policy that minimizes the OT distance to the risk distribution also maximizes the expected number of visits to states close to the risk distribution. This formally establishes the desired result.

\noindent{\textbf{Implications.}} Theorem 4 validates the intuitive idea that when a policy reduces its OT distance to a desired risk distribution, it inherently increases its visits to states that align closely with that risk distribution. This suggests a natural mechanism for RL agents to exhibit safer behaviors: by minimizing the OT distance to a target risk profile, the agent is steered towards states that are deemed safer.

\section{Discussion}

The integration of risk-aware RL with OT has inaugurated a new direction in how we approach risk management in uncertain environments. The OT framework offers a panoramic and robust risk measurement that captures the entire distribution of states, transcending traditional risk metrics that often rely on isolated statistics. Such an approach ensures a more comprehensive understanding of risk while preserving the dynamism of states. Moreover, the inherent adaptability of our method permits the agent to adjust its risk perception based on specific contexts or tasks. However, the very richness of OT also brings forth challenges, especially regarding computational complexity in high-dimensional environments, potentially hampering its real-time utility in certain domains. Additionally, the efficacy of the approach is tightly coupled with the choice of risk distribution, which, though flexible, may introduce complexities in decision-making. As we look towards the horizon, it becomes imperative to address these computational challenges, possibly through algorithmic innovations that marry efficiency with the core benefits of OT. Empirical validations across a plethora of RL scenarios stand paramount to not only corroborate our theoretical insights but also to refine the approach for varied applications.

\section{Conclusions}

In this work, we have proposed a formulation for risk-aware RL grounded in the mathematical framework of OT theory. This approach seeks to incorporate risk considerations into the heart of RL algorithms. We have provided a series of theorems that clarify the relationship between risk distributions, optimal value functions, and policy behaviors. These theorems highlight the trade-offs between maximizing rewards and safeguarding against risks. Importantly, our theorems demonstrate that minimizing OT costs can yield policies that are not only reward-optimal but also intrinsically safer.

\bibliographystyle{IEEEtran}
\bibliography{ref}

\end{document}